\documentclass[letterpaper, 10 pt, journal, twoside]{IEEEtran}  

\IEEEoverridecommandlockouts 

\usepackage{graphics} 
\usepackage{epsfig} 
\usepackage{mathptmx} 
\usepackage{times}
\usepackage{amsmath} 
\usepackage{amssymb}  
\usepackage{float}
\usepackage{array}
\usepackage{mdwmath}
\usepackage{mdwtab}
\usepackage{todonotes}
\usepackage[switch, pagewise]{lineno}
\usepackage{url}
\usepackage{paralist}
\usepackage{caption}
\usepackage{subcaption}
\usepackage[version-1-compatibility]{siunitx}
\usepackage{booktabs}
\newcommand{\etal}{~et al.~}
\usepackage{hyperref}
\usepackage{pifont}

\usepackage{pdfpages}

\begin{document}
\title{Fast and Continuous Foothold Adaptation for Dynamic Locomotion through 
CNNs}

\markboth{IEEE Robotics and Automation Letters. Preprint Version. Accepted 
January, 2019}
{Villarreal \MakeLowercase{\textit{et al.}}: Fast Foothold Adaptation for 
Dynamic Locomotion through CNN}

\author{Octavio Villarreal$^{1}$, Victor Barasuol$^{1}$, Marco 
		Camurri$^{1,4}$,  Luca Franceschi$^{2}$, \\Michele Focchi$^{1}$, 
		Massimiliano Pontil$^{2}$, Darwin G. Caldwell$^{3}$ and Claudio 
		Semini$^{1}$%
		\thanks{Manuscript received: September, 10, 2018; Revised December, 16, 
		2018; 
		Accepted January, 25, 2019.}
		\thanks{This paper was recommended for publication by Editor Asfour, 
		Tamim upon evaluation of the Associate Editor and Reviewers' 
		comments. 
		This work was supported by Istituto Italiano di Tecnologia.}
		\thanks{Octavio Villarreal, Victor Barasuol, Michele Focchi and 
		Claudio Semini are 
		with the Dynamic Legged Systems lab, Istituto Italiano di Tecnologia,
			Via Morego 30, 16163 Genoa, Italy.
			{\tt\small firstname.lastname@iit.it}}%
		\thanks{Luca Franceschi and Massimiliano 
			Pontil are with the Computational Statistics and Machine Learning, 
		Istituto 
		Italiano di Tecnologia, Via Morego 30, 16163 Genoa, Italy.
			{\tt\small firstname.lastname@iit.it}}
		\thanks{Darwin G. Caldwell is with the Department of Advanced 
		Robotics, Istituto Italiano di 
		Tecnologia, Via Morego 30, 16163 Genoa, Italy.
			{\tt\small darwin.caldwell@iit.it}}
		\thanks{Marco Camurri is with the  Dynamic Legged Systems lab, 
		Istituto Italiano di Tecnologia,
			Via Morego 30, 16163 Genoa, Italy and the Oxford Robotics 
			Institute, 
			University of Oxford, 23 Banbury 
		Rd, OX2 6NN Oxford, UK.
			{\tt\small mcamurri@robots.ox.ac.uk}}
		\thanks{Digital Object Identifier (DOI): see top of this page.}
}
\null
\includepdf[pages=-]{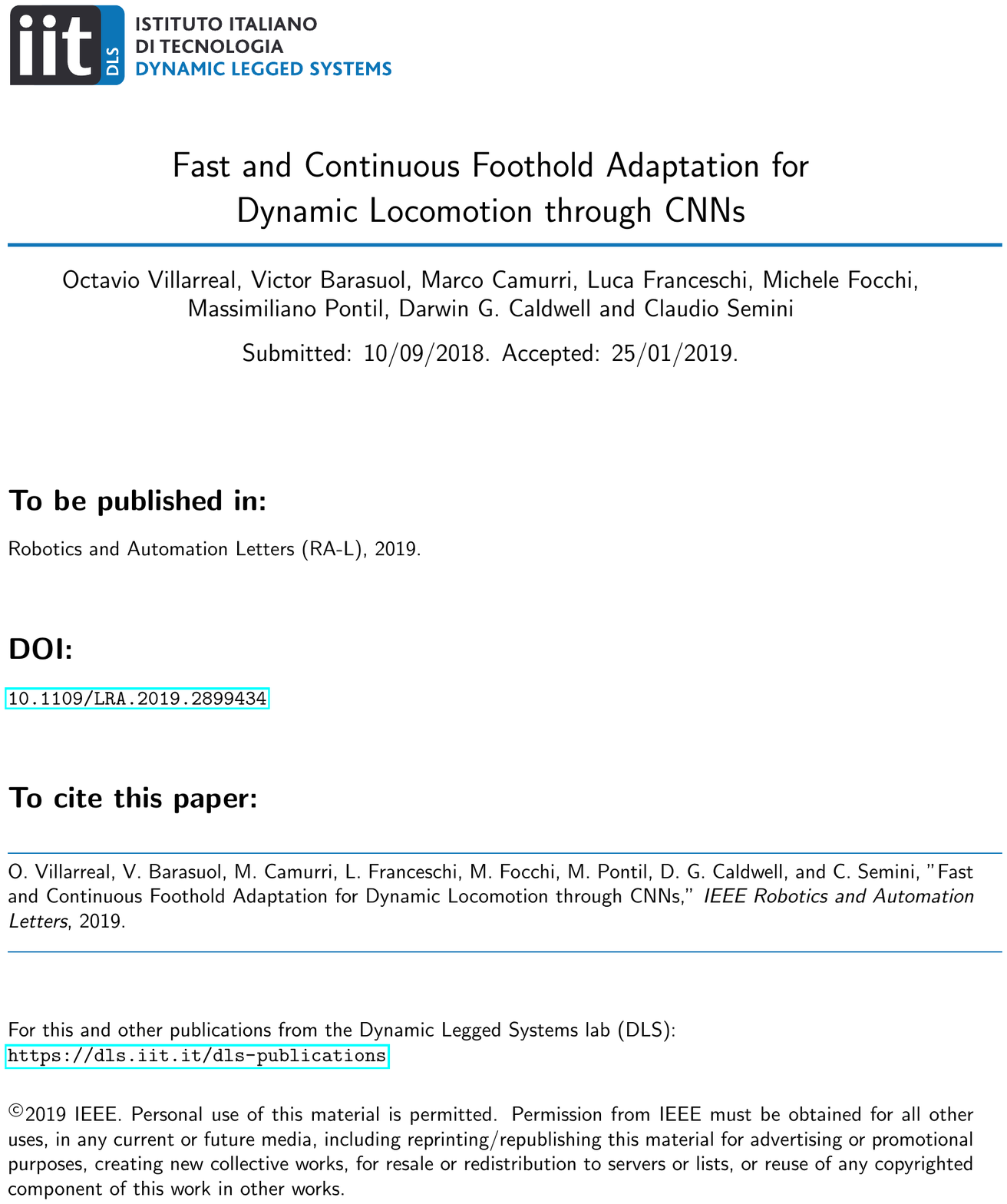}
\pagenumbering{arabic}
\maketitle
\begin{abstract}
Legged robots can outperform wheeled machines for most navigation tasks across
unknown and rough terrains. For such tasks, visual feedback is a fundamental
asset to provide robots with terrain-awareness. However, robust dynamic
locomotion on difficult terrains with real-time performance guarantees remains
a challenge. 
We present here a
real-time, dynamic foothold adaptation strategy based on visual feedback. Our
method adjusts the landing position of the feet in a fully reactive manner,
using only on-board computers and sensors. The correction is computed and
executed continuously along the swing phase trajectory of each leg. To
efficiently adapt the landing position, we implement a self-supervised foothold classifier based on a
Convolutional Neural Network (CNN). 
Our method results in an up to 200 times faster computation
with respect to the full-blown heuristics. Our goal is to react to
visual stimuli from the environment, bridging the gap between blind
reactive locomotion and purely vision-based planning strategies. We
assess the performance of our
method on the dynamic quadruped robot HyQ, executing static and dynamic gaits 
(at speeds up to 0.5 m/s) in
both simulated and real scenarios; the benefit of safe foothold adaptation is
clearly demonstrated by the overall robot behavior.
\end{abstract}
\begin{IEEEkeywords}
Legged Robots; Reactive and Sensor-Based Planning; Deep Learning in Robotics 
and Automation.
\end{IEEEkeywords}

\IEEEpeerreviewmaketitle

\section{Introduction}
\IEEEPARstart{L}{egged} platforms have recently gained increasing 
attention, motivated by the versatility that these machines can offer over a 
wide variety of terrain and tasks. Quadrupeds in particular are 
able to perform robust
locomotion in the form of statically
\cite{focchi2016,hirose2000} and dynamically
\cite{barasuol13icra,park17ijrr} stable gaits. In parallel, sensor fusion
techniques have evolved to overcome the
harsh conditions typical of field operations on legged machines
 \cite{nobili_camurri2017rss}, to  provide effective pose and velocity
estimates for planning \cite{mastalli15tepra}, control, and mapping
\cite{fankhauser14clawar,camurri15mfi}.

Despite this progress, a real-time safe, computationally
efficient way to use 3D visual feedback in dynamic legged locomotion has not been 
presented yet. The
challenge lies on the high-density nature of visual information, which makes it
hard to meet the fast response requirement for control actions at dynamic
locomotion regimes.

The use of exteroceptive feedback in locomotion has been successfully demonstrated in the
past, yet most approaches are
limited by the dependency on external motion capture \cite{kalakrishnan09iros},
the execution of precomputed trajectories in open-loop
\cite{aceituno2017ral,wermelinger16iros}, and/or to statically stable gaits
\cite{fankhauser18icra}.

In this paper, we focus on difficult scenarios, where the presence of 
disturbances and rough terrain may lead to deadlocks (e.g., getting stuck with 
an
obstacle). Furthermore, we want to
perform this task during dynamic locomotion. To this end, we
propose a real-time foothold adaptation strategy that uses
only on-board sensing and computation to execute reactive corrections while the
robot navigates through rough terrains.

The strategy proposed here is based
on our previous work \cite{barasuol15iros}, where we implemented a supervised
learning algorithm based on expert demonstration and a logistic regression
model.  
\begin{figure} [b]
	\centering
	\includegraphics[width=1\columnwidth]{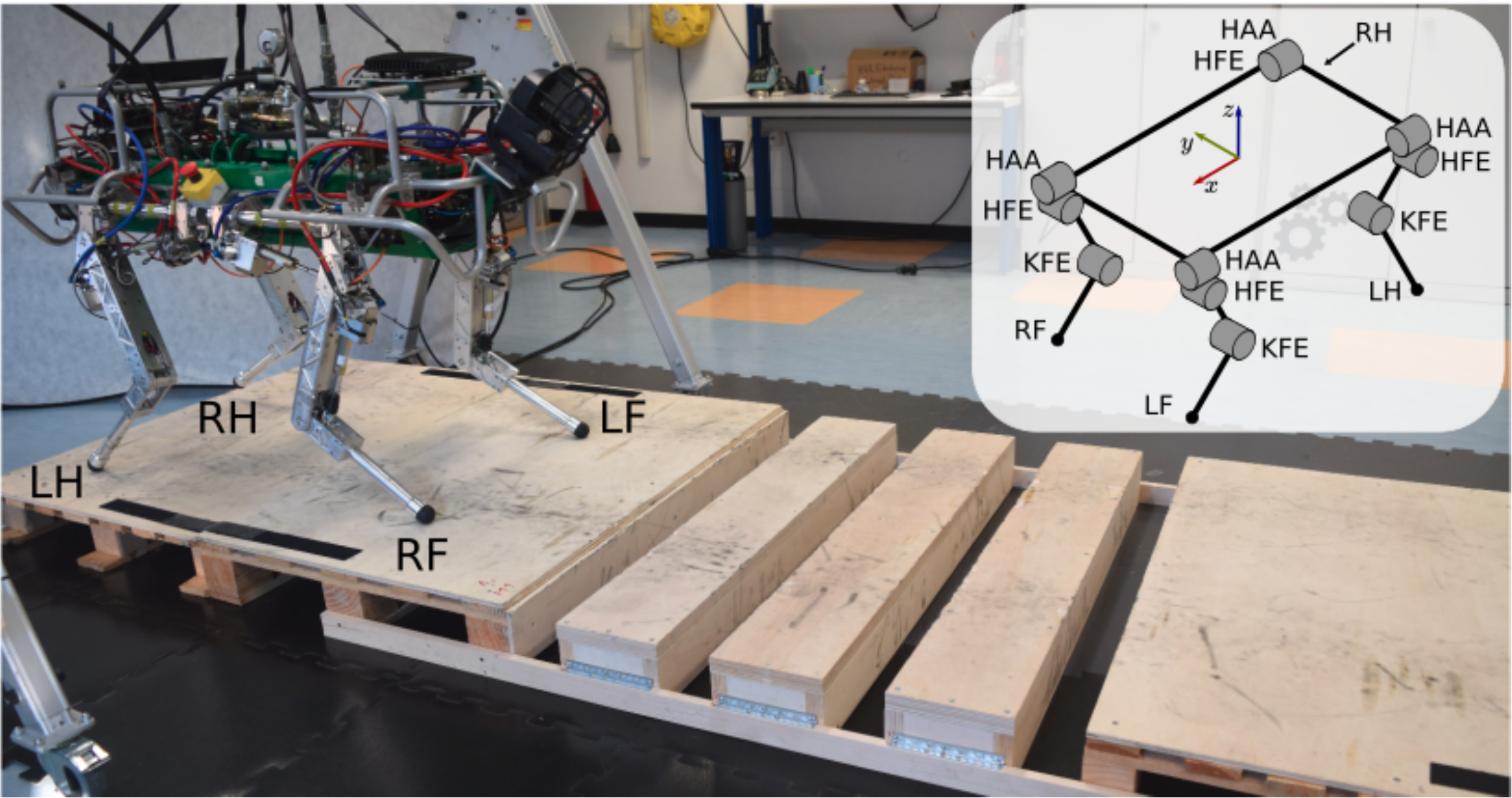}
	\caption{\small{HyQ robot positioned on the terrain
		template used for experimental test. 
		The naming conventions
		for the robot joints, axes and legs are seen in the top-right image. The
		legs are identified as LF (left-front), RF (right-front), LH (left-hind)
		and RH (right-hind). The leg joints are identified as HAA
		(hip adduction-abduction), HFE (hip flexion-extension) and KFE (knee 
		flexion-extension).}}
	\label{fig:hyq_overview}
\end{figure}
Our strategy does not rely on visual information only, but rather
acts as an interface for the reactive layer of our locomotion controller
\cite{barasuol13icra}. The idea is to enhance such controller with reliable
feedback obtained from exteroceptive sensing, to increase the traversability
of difficult environments. 
Preliminary simulation results of the strategy here proposed were 
presented in the 
short workshop 
paper \cite{villarreal18iros}.

\noindent The contributions of this paper are summarized as follows:
\begin{enumerate}
\item To the best of our knowledge, the proposed approach is the first to 
achieve reactive and real-time
obstacle negotiation for dynamic gait locomotion, with
full on-board computation (control, state estimation and mapping). The fast 
speed of computation and execution allows the robot to adapt the foot trajectory
\textit{continuously} during the swing motion of the legs, and grants
the robot the capability to react favorably against disturbances applied
on the trunk at any given time;
\item We improved our previous work \cite{barasuol15iros} in terms of 
\textit{autonomy of 
training}. We replaced the
human expert with a heuristic algorithm that generates the ground
truth from the terrain morphology, kinematic configurations, foot and leg
collisions. This makes the
approach self-supervised, and allows to generate more (potentially
unlimited) training samples
(3300 in \cite{barasuol15iros} vs. 17844 in this work). We also increased the
possible outputs (landing positions) from 9 in \cite{barasuol15iros} to 225;
\item We improved the \textit{prediction} model with respect to 
\cite{barasuol15iros} by 
replacing the
logistic classifier with a Convolutional Neural Network (CNN), allowing for 
more complex inputs (i.e., more
difficult obstacles) to be processed successfully. To the best of our
knowledge, this is the
first time a CNN is used to learn foothold corrections in legged
locomotion. CNNs are
very effective for image processing
\cite{krizhevsky2012imagenet,goodfellow2016deep}, and are here efficiently implemented to incorporate the knowledge of an effective (yet computationally
expensive) heuristic algorithm.  This is achieved through low-dimensional
parameterization and a carefully balanced network architecture.
\end{enumerate}

The remainder of this paper is organized as follows: Section
\ref{section:related_work} summarizes
the work related to our proposed strategy; Section \ref{section:system_overview}
provides a description of the HyQ platform used to test the proposed 
strategy;
Section \ref{section:visual_adaptation} describes the
methods to select a safe foothold; simulation and experimental results are shown in Section
\ref{section:results}; finally, the conclusions and future work are
presented in Section \ref{section:conclusions}.
\section{Related Work}
\label{section:related_work}
Kolter\etal \cite{kolter08icra} have provided one of the first applications of
terrain awareness to enhance the traversing capabilities of a quadruped
robot. To do so, collision
probability maps and  heightmaps collected a priori are used to train
a Hierarchical Apprenticeship Learning algorithm, to select the best footholds
in accordance to an expert user. 

A similar approach was taken by Kalakrishnan\etal \cite{kalakrishnan09iros}. In
contrast to \cite{kolter08icra}, visual feedback was discretized using
templates,
i.e., portions of terrain in the vicinity of a foothold. With a
learning regression method based on expert user selection, a target foothold is
associated to each template. The authors have incorporated the classification
algorithm into a locomotion planner and demonstrated its validity on
the robot LittleDog, traversing highly unstructured terrains.

Both approaches have proved to be powerful, but they rely on external
motion capture systems, reducing their field of application to
controlled and calibrated environments.
In contrast, Belter \etal \cite{belter11jfr} used an on-board laser scanner to
collect an elevation map of the terrain.  
Their method searches for useful clues related to the foothold
placement,
and selects the ones with minimal slippage. The optimal footholds are
learned in an unsupervised fashion, inside a simulated environment.

Despite their ability to perform locomotion tasks with on-board sensors only,
most of the vision-based foothold selection strategies involve slow motions, mainly to provide enough time to complete the most costly operations such as image
processing and optimization. An exception was shown by Wahrmann\etal
\cite{wahrmann16aim}, where the acquisition of swept-sphere-volumes allowed the
biped robot Lola to avoid obstacles while moving, with no
prior information about the environment. 
Nevertheless,
this strategy was mainly demonstrated for single obstacle avoidance and 
self 
collision, and not  
for rough terrain.

Our previous work \cite{barasuol15iros} is similar to the template-based foothold correction of
\cite{kalakrishnan09iros}, but it differs due to its implementation in a fully reactive fashion.
Heightmaps around the nominal footholds are evaluated to generate continuous
motion corrections for the \textit{Reactive Controller Framework} (RCF)
\cite{barasuol13icra}. The corrections are learned from expert demonstration
using a Logistic Regression classifier.

More recently, Fankhauser \etal \cite{fankhauser18icra} presented a
perception-based statically stable motion planner for the quadruped robot
ANYmal. For each footstep, the algorithm generates a
foothold (upon
rejection of unsafe and kinematically unfeasible solutions), a collision free
foot trajectory, and a body pose. The work here presented differs from 
this because it can
deal with dynamic gaits (e.g., a trotting gait), it accounts for the leg
collisions when generating collision-free trajectories for the foot, and it
can deal with external disturbances during the whole locomotion stride.
\section{System Overview}
\label{section:system_overview}
The quadruped robot HyQ \cite{semini11HyQdesignJSCE} (Fig.
\ref{fig:hyq_overview}) is a hydraulically actuated, versatile research
platform. It weighs
\SI{90}{\kilo\gram}, is \SI{1}{\meter} long and \SI{1}{\meter} tall. Each
 leg has 3 Degrees-of-Freedom (DoF): a Hip joint for
Abduction/Adduction (HAA, actuated by a rotary hydraulic motor), a Hip joint
for Flexion/Extension (HFE), and a Knee joint for Flexion/Extension (KFE).
The latter two joints are actuated by hydraulic cylinders.
\paragraph{Sensors}
HyQ is equipped with a variety of proprioceptive sensors (for a detailed
reference, see
\cite{camurri2017phdthesis}), including: a tactical-grade IMU (KVH 1775), 
8
loadcells (located in all the HFE and KFE joints) and 4 torque sensors (located
at the motors of the HAA joints).
Each joint's position is measured with a high-resolution optical
encoder.
These sensors are synchronized by the EtherCAT network, 
with a maximum latency of \SI{1}{\milli\second}.

\begin{figure}
		\centering
			\includegraphics[width=\columnwidth]{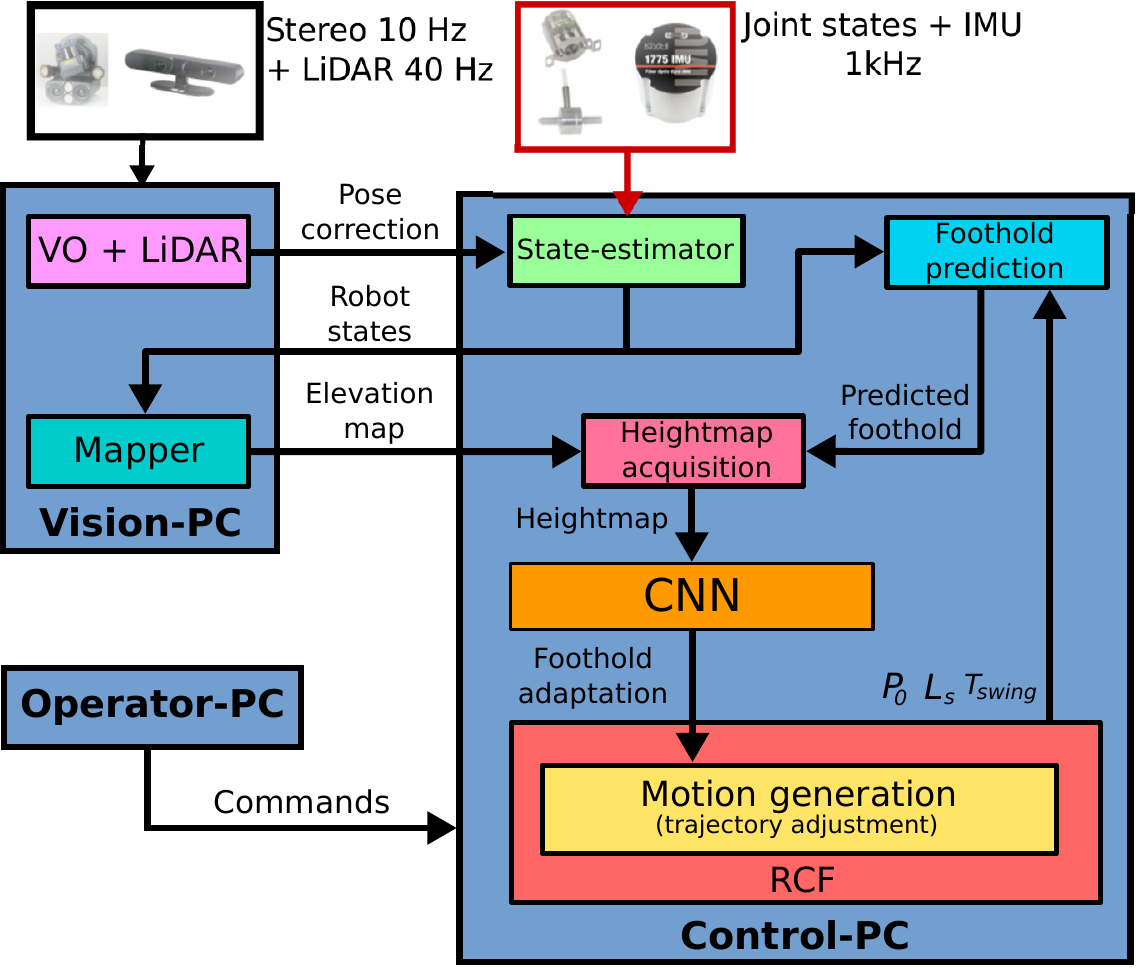}
		\caption{\small{Schematic drawing of our software
		architecture. The proprioceptive core of the state  estimator (green
		box) runs on the Control-PC, while the low frequency updates from Visual
		Odometry and LiDAR scan matching are received from the Vision-PC (see
		\cite{nobili_camurri2017rss}). The foothold prediction, the 
		heightmap acquisition and the CNN-based foothold adaptation are 
		executed 
		inside the Control-PC. The CNN sends the adaptation commands to the RCF 
		motion 
		generation 
		module.}}
		\label{fig:hyq_software}
\end{figure}

Exteroceptive sensors include: an ASUS Xtion RGB-D sensor for mapping; a
 Multisense SL for pose estimation (Visual Odometry (VO) and LiDAR scan
matching). 
The main sensor characteristics are summarized in \cite{nobili_camurri2017rss}.
\paragraph{Hardware/Software architecture}
HyQ is equipped with a Control-PC, running a real-time Linux kernel, and
a Vision-PC, running a regular Linux kernel. The two computers are
synchronized by means of an NTP server. The first computer executes the robot
control commands in a real-time environment, as well as the Extended Kalman
Filter
state estimator in a non-real-time thread. The Vision-PC collects the
exteroceptive inputs, computes the visual odometry and
ICP-based scan matching (as described in \cite{nobili_camurri2017rss}), and
delivers to the Control-PC an elevation map surrounding the robot 
(see Fig. \ref{fig:hyq_software}). In case of failure of the
Vision-PC, the controller would still be able to operate blindly
with a smooth but drifting pose estimate. 
\section{Visual Foothold Adaptation for 
Locomotion}\label{section:visual_adaptation}
In this section, we explain our strategy to deal with 
computationally demanding visual information. We seek to embed 
domain 
knowledge
from legged locomotion into a CNN-based learning algorithm. This strategy
is primarily applied to the trotting motions from the RCF 
\cite{barasuol13icra}. For the sake of generality, we
have also applied our strategy to the \textit{haptic crawl} of 
\cite{focchi2016}. The
key elements of our strategy are:
\begin{inparaenum}
	\item prediction of the next foothold for each leg;
	\item acquisition of heightmap information in the vicinity of the 
	next foothold;
	\item foothold evaluation based on kinematics and terrain roughness;
	\item training and learning based on a CNN;
	\item feet trajectory adjustment for foothold adaptation.
\end{inparaenum}
These elements are explained in detail next.
\vspace{-0.3cm}
\subsection{Prediction of the nominal 
foothold}\label{section:foothold_prediction}
With foothold prediction we indicate the estimation of the
landing position of a foot during the leg's swing phase. This quantity, expressed
in the world frame, is hereafter defined as \textit{nominal
foothold}. The computation of a nominal foothold differs significantly 
depending on the motion of the trunk. 

Some crawl gait implementations do not move the trunk during the swing 
phase 
motion of the legs (e.g., our haptic crawl \cite{focchi2016}). The nominal
foothold can be
computed at lift-off according to the desired direction of motion. Therefore, 
the only source of error between the nominal and
the actual foothold comes from foot trajectory tracking.

On the other hand, in gaits that yield motion of the trunk during swing phase 
(e.g., a diagonal trot), the nominal foothold has to account for
the trunk velocity in addition to the foot trajectory tracking. Hence there are
two sources of uncertainty: trajectory tracking and trunk state estimation
(position and velocity).

In the RCF, the foot swing trajectory is described by a half ellipse, where the
major axis corresponds to the step length. To compute the nominal foothold, we
use the following approximation:
\begin{equation}
P_n = P_0 + \frac{L_s}{2} + (T_{swing} - t_{swing})\dot{X}_b,
\label{eq:predicted_foothold}
\end{equation}
where $P_n$ is the nominal foothold position in world coordinates,
$P_0$ is the position of the center of the ellipse at lift-off in world 
coordinates, $L_s$
is the step length vector, $T_{swing}$ is the swing
period (defined by the duty factor $D_f$ and the step frequency $f_s$), $t_{swing}$
is the time elapsed from the latest lift-off event to the touchdown event, and $\dot{X}_b$ is the trunk velocity. Intuitively, the second term on the
right hand side of \eqref{eq:predicted_foothold} is the distance covered by the 
leg due to the leg trajectory execution, while the third term is the distance
travelled by the trunk, assuming that $\dot{X}_b$ is constant over the rest
of the swing phase $T_{swing} - t_{swing}$.

In \eqref{eq:predicted_foothold}, $P_0$ and $\dot{X}_b$ are taken from the state
estimator and are therefore affected by uncertainty (see
\cite{nobili_camurri2017rss}). To understand the effects of this uncertainty,
we conducted a series of preliminary experiments with the robot trotting on
flat terrain. A comparison between the actual foot landing position and the
predicted one along the swing phase from \eqref{eq:predicted_foothold} showed
an average error of approximately \SI{3}{\centi \meter}.
\vspace{-0.3cm}
\subsection{Foothold heightmap}\label{section:foothold_heightmap}
We define as \emph{foothold heightmap} (or simply \emph{heightmap}) a
squared, bidimensional and discrete representation of the terrain where
each pixel describes the height of a certain area.
The heightmap 
is obtained considering its center as the nominal foothold and oriented with 
respect to the \textit{Horizontal Frame} of the robot, which is a frame 
whose 
origin coincides with the body frame, with the $xy$ plane always 
perpendicular to the gravity vector (for a detailed 
explanation of the Horizontal Frame, see \cite{barasuol13icra}). Since 
we are considering robots with point-feet, we define the foothold as a 3-D space
position.

Given a nominal foothold, the heightmap around it can be easily extracted from 
the
elevation map computed \textit{on-board} by the Vision-PC (see Section
\ref{section:system_overview}). We obtain this elevation map using the Grid Map 
interface from \cite{Fankhauser2016GridMapLibrary}. The heightmap is then 
analysed to adapt the
landing position of the feet and avoid unsafe motions (see
Section \ref{section:foothold_adaptation}).

\begin{figure}
	\centering
	\begin{subfigure}[t]{0.56\linewidth}
		\centering
		\includegraphics[width=1\linewidth]{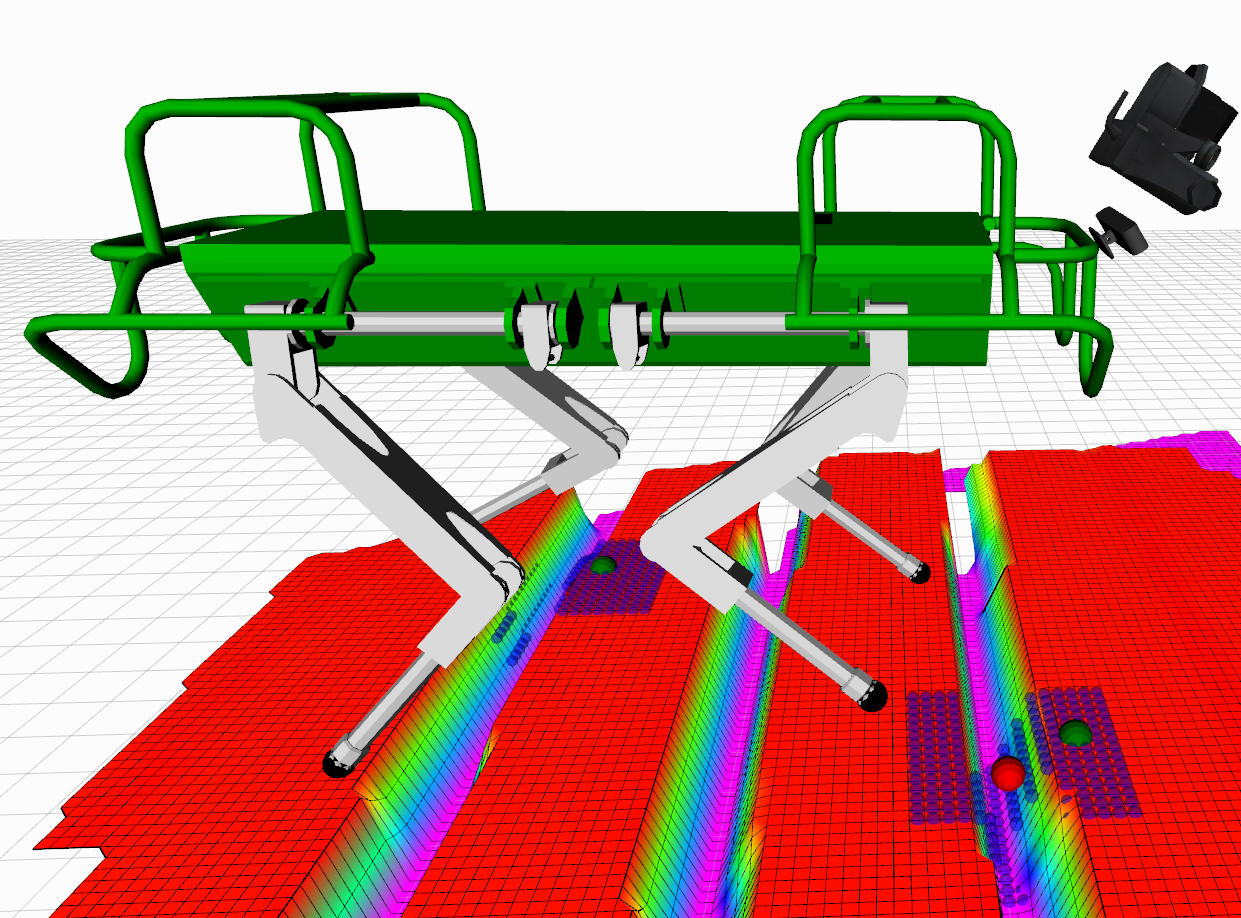}
	\end{subfigure}
	\quad
	\begin{subfigure}[t]{0.38\linewidth}
		\centering
		\includegraphics[width=0.99\linewidth]{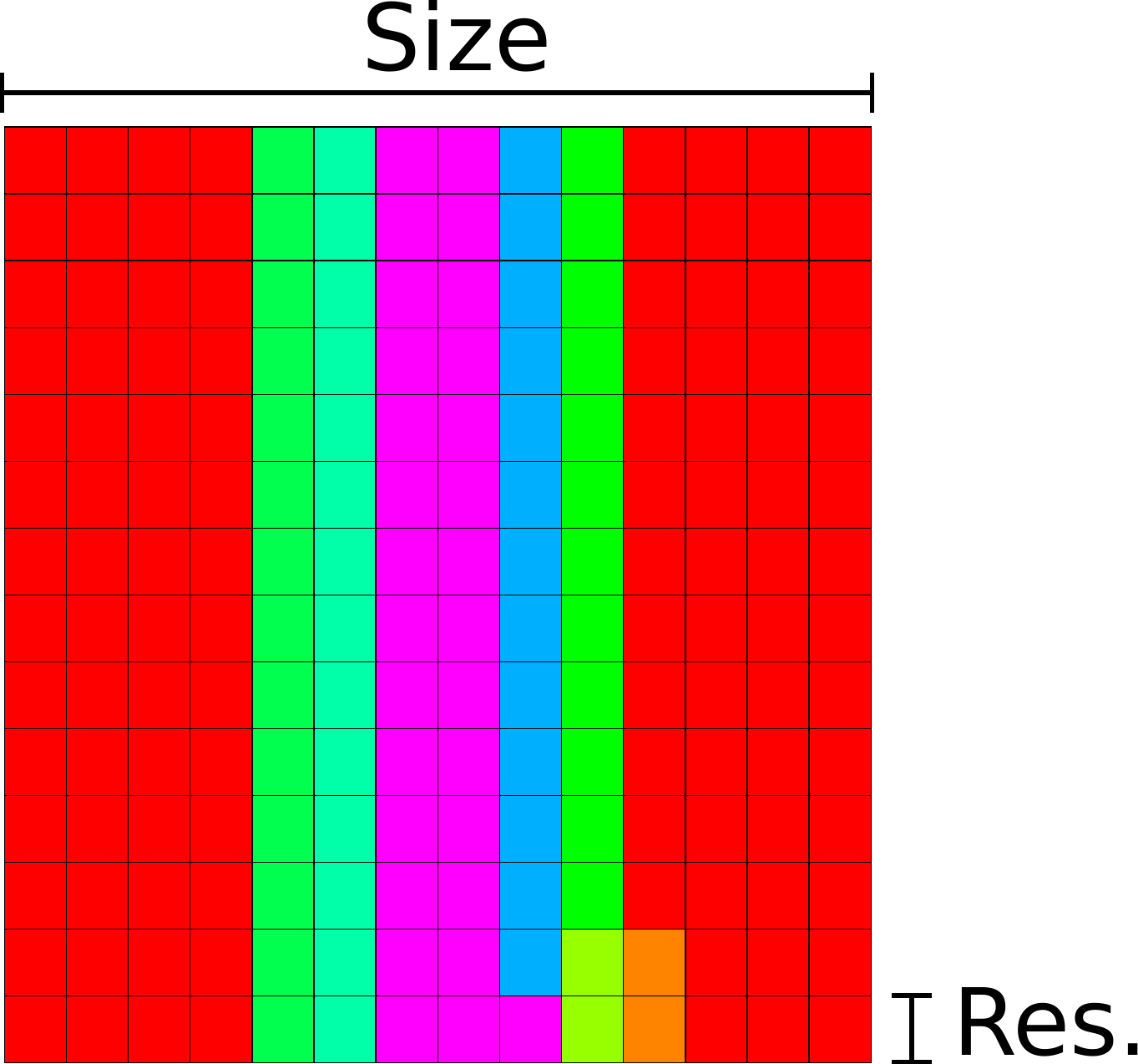}
	\end{subfigure}
	\caption{\small{Simulation of HyQ trotting while acquiring an elevation map of the terrain (represented in false colors) using the RGB-D
	sensor (\textit{left}). The heightmaps used
	as inputs to the neural network are shown as squares comprised by the blue 
	spheres, where each sphere is a potential foothold. The
	nominal foothold is represented by the red
	sphere,
	 while the corrected foothold is represented by the green sphere. On the 
	 right,
	 the heightmap corresponding to the right front leg is shown in false 
	 colors. Size and resolution are also indicated. 
	 The 
	 dark red intensities correspond to the maximum height values, whereas the 
	 dark 
	 magenta intensities correspond to the minimum ones.   
	 }}\label{fig:heightmaps}
	 \vspace{-0.35cm}
\end{figure}

The heightmap is parametrized by size and
resolution. These two parameters are depicted in \ref{fig:heightmaps}.
Both parameters are
the result of a compromise between computational expense and task requirements.
We want to avoid processing large amounts of data, while retaining a level
of detail that is meaningful for the 
given task.
A detailed
discussion on appropriate parameter selection can be
found in \cite{barasuol15iros}. Fig. \ref{fig:heightmaps} shows an example of
a heightmap. Each pixel of the image corresponds to a possible foothold.
When dimensioning the heightmap, we would like to avoid blind spots
(i.e., empty areas between two consecutive foothold heightmaps). We opt for 
$30\times 30$ \SI{}{\centi\meter} heightmap size, with a resolution of 
\SI{2}{\centi\meter\squared} for each pixel (heightmap of $15\times 15$ 
pixels).

Since drift-free and real-time mapping is still
an open issue for dynamic motions, we analyze the degree of uncertainty
coming from the map and consider a safety margin to
avoid dangerous drifted map locations (see Section
\ref{section:foothold_adaptation}).
\vspace{-0.3cm}
\subsection{Foothold Adaptation Heuristic Criteria}
\label{section:foothold_adaptation}
In this section, we describe the heuristic algorithm used to train
automatically the CNN (described in Section \ref{section:learning}). The 
algorithm evaluates
each foothold inside the heightmap according to the following criteria:
\paragraph{Kinematics}if a foothold is outside the workspace of the
robot leg, the pixel is discarded.
\paragraph{Terrain Roughness} for each heightmap pixel, 
we compute the mean 
and the standard deviation of the slope relative to its neighborhood.
We define a specific threshold for the 
sum of the standard deviation and the mean of the slope, according to the 
foot radius. The footholds 
with values above this 
threshold are discarded. 
\paragraph{Uncertainty Margin} to account for uncertainty, footholds 
within a 
certain radius that are deemed as unsafe according to the terrain roughness, 
are discarded. This also implicitly prevents from stepping close to
obstacle edges. 

\begin{figure}
		\centering
		\begin{subfigure}[t]{0.52\linewidth}
			\centering
			\includegraphics[width=1\linewidth]{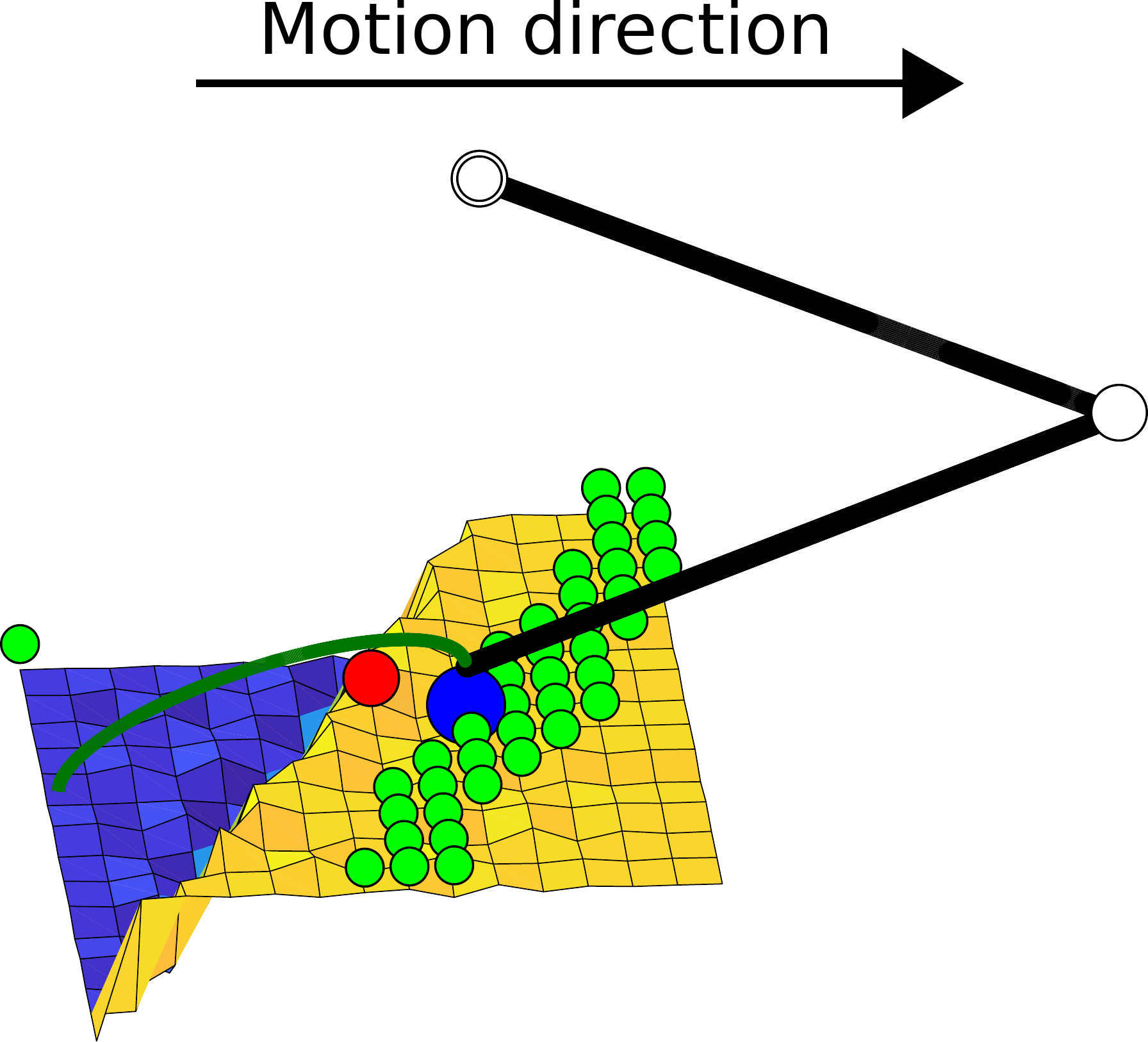}
		\end{subfigure}
		\quad
		\begin{subfigure}[t]{0.41\linewidth}
			\centering
			\includegraphics[width=0.99\linewidth]{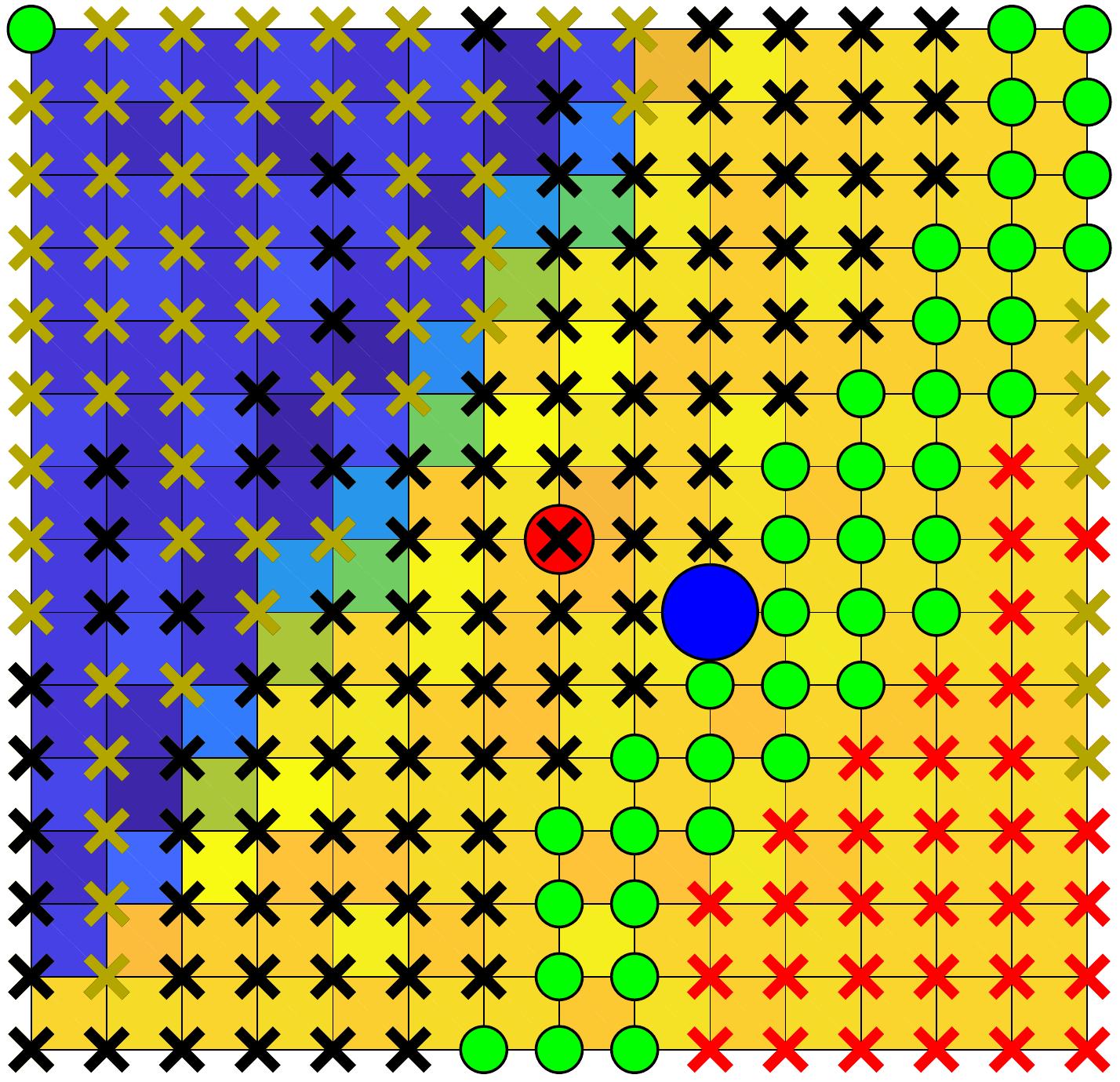}
		\end{subfigure}
	\caption{\small{Example of evaluated foothold heightmap.
	The left image shows a hind leg and the adapted trajectory. The solid 
	black lines are the
	upper and
	lower limbs of
	the leg and the dark green solid line represents the foot trajectory. 
	The red sphere is the nominal foothold, the
	green spheres are feasible footholds, and the blue one is
	the optimal according to the heuristics. A top view of the same 
	heightmap is shown on the right. In addition to the feasible 
	footholds, 
	crosses correspond to the discarded footholds due to: the uncertainty 
	margin and terrain roughness (black), foot frontal collision (red) 
	and shin collision (light brown).}}\label{fig:correction_examples}
\end{figure}

We experimentally identified an uncertainty of
\SI{3}{\centi \meter} around a nominal foothold, related to errors in the 
foothold prediction due to the trunk state estimation. In addition, a short 
term map drift of \SI{2.5}{\centi\meter} (mainly due to pose estimation drift)
has also been observed after traversing a distance of approximately
\SI{2}{\meter}. 
\paragraph{Frontal Collision} for each pixel, we evaluate potential foot 
frontal
collisions along the corresponding trajectory from the lift-off location. In
Fig. \ref{fig:correction_examples}, the pixels marked with a red
\ding{54} symbol correspond to discarded footholds, due to frontal collisions.
\paragraph{Leg Collision} similarly to frontal collisions,
we evaluate the intersection between the terrain and both leg limbs throughout
the whole step cycle (i.e., stance and swing phases).
The discarded footholds due to leg collision are shown in Fig.
\ref{fig:correction_examples} with a light brown \ding{54}.
\paragraph{Distance to nominal foothold} given all valid footholds upon
evaluation of the previous criteria, the algorithm chooses as optimal the
one closest to the
nominal foothold. This minimizes the deviation from the original
trajectory.

Let $x\in\mathbb{R}^{n\times n}$ be an input heightmap. We 
denote\footnote{Herein, $\mathbb{Z}_2 = \{0,1\}$, where 0 corresponds to an 
unsafe foothold and 1 to a safe foothold.} 
$g_i(x):\mathbb{R}^{n\times n}\to\mathbb{Z}_2^{n\times n}$ as a mapping that 
takes a heightmap $x$ as input and outputs a matrix of binary values 
indicating 
the elements in $x$ that correspond to safe footholds according to 
one of the previous criteria. We define four mappings $g_k$, 
$g_t$, $g_f$ and $g_l$, corresponding to: kinematics; terrain roughness and 
uncertainty margin; frontal collision; and leg collision, respectively. A 
\textit{feasible foothold} is defined as a foothold that is 
deemed as safe according to all four mappings. Furthermore, we define 
$g(x)$ as 
\begin{equation}
\label{eq:mapping}
	g(x) = g_k(x) \wedge g_t(x) \wedge g_f(x) \wedge g_l(x),
\end{equation}
where the operator $\wedge$ represents a coefficient-wise logical 
\textit{AND}. The elements that are equal to $1$ of the matrix coming out of 
$g(x)$ correspond to 
feasible footholds. The heuristic mapping $h:\mathbb{R}^{n\times 
n}\to\{0,1\}^m$, computes the feasible 
footholds according to $g(x)$ and outputs a ``one-hot vector'' that represents 
the optimal landing point, as the one with the smallest Euclidean 
distance to the nominal foothold. 

Fig. \ref{fig:correction_examples} shows the feasible footholds as green 
spheres, and the optimal foothold with
a blue sphere.
\vspace{-0.3cm}
\subsection{CNN Training}
\label{section:learning}
We decided to use a CNN as model for predicting the outputs of the heuristics, 
since it has better predictive capabilities (yet low computational 
requirements) than the Logistic Regressor we used in \cite{barasuol15iros}. Our 
input to the CNN are matrices of $15\times15$ corresponding to foothold heightmaps. 
As output, we obtain a foothold corresponding to one of the labelled 
pixels that 
represent a landing position, as depicted in the right image of Fig. 
\ref{fig:correction_examples}. 

To generate the training set, heightmaps can be obtained from three 
different data sources: simulation, artificial generation or experiments. In 
this paper, we only use simulated and artificially generated data. 
Simulated data refers to heightmaps collected from our simulation 
environment in Gazebo, including sensors and the robot dynamics. 
Artificial data 
corresponds to heightmaps generated with no physical or sensor 
data coming from simulation. 
We define the elements of the
$15\times15$ matrices in MATLAB to represent bars, gaps and 
steps of 
different lengths, heights and at various orientations.

Let again $ x\in\mathbb{R}^{n\times n}$ be an input heightmap. Note that 
in our 
setting $n=15$.
We denote by $f_w:\mathbb{R}^{n\times n}\to[0,1]^m$ the CNN classifier parametrized by a
weight vector $w\in\mathbb{R}^d$, where $d = 8238$ (number of parameters 
of the
CNN) and $m=226$ (number of potential 
footholds).
Using the heuristic mapping $h(x)$ outlined in Section 
\ref{section:foothold_adaptation}, we build a dataset of
$N$ labeled examples $D=(x_i, h(x_i))_{i=1}^N$ from simulated and 
artificially 
generated data. To improve the robustness against noisy heightmaps, the 
training set is corrupted by
zero-mean Gaussian noise with a standard deviation of \SI{3}{\centi\meter}.

The network is trained to minimize the cross-entropy loss on $D$. Namely, 
we
approximately
solve the following optimization problem
\begin{equation}
\min_w - \frac{1}{N}\sum_{i=1}^N h(x_i) \circ \log f_w(x_i),
\end{equation}
by stochastic gradient descent, where $\circ$ denotes element-wise product.

\begin{figure}
	\centering
	\includegraphics[width=\linewidth]{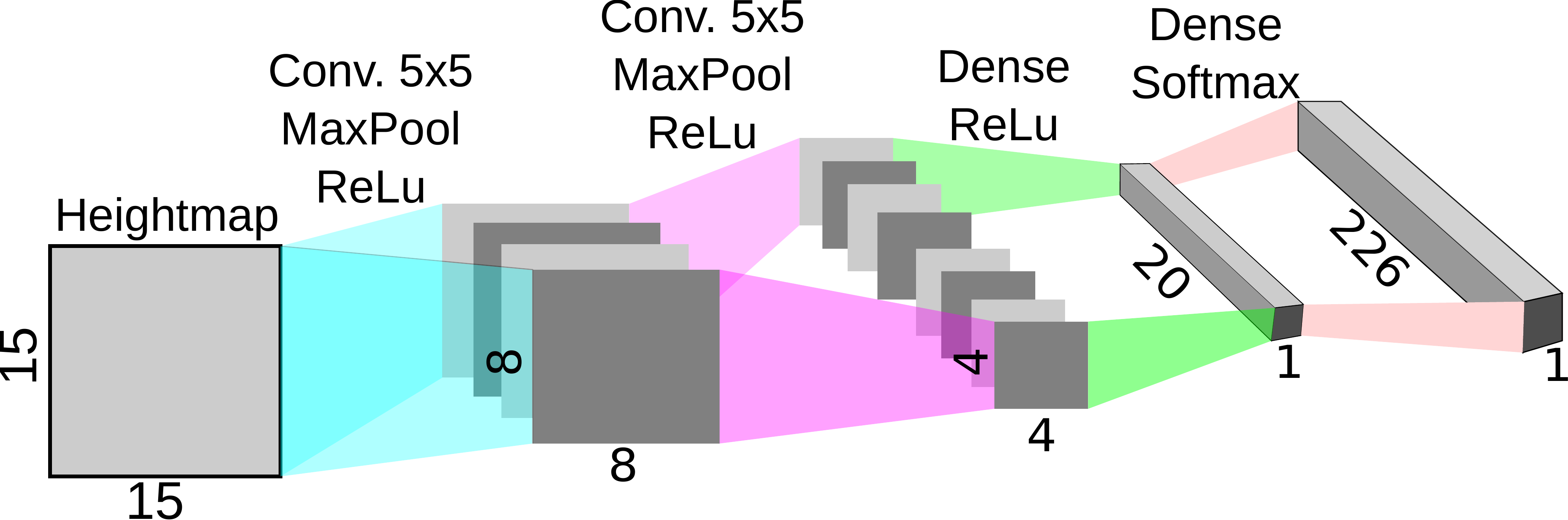}
	\caption{\small{Architecture of the convolutional neural network. The input 
	is a $15\times15$ matrix
	corresponding to a heightmap around the vicinity of a foothold. There
	are 2 convolutional layers, the first one performs convolution of 
	the heightmap with 4 kernels and
	the second takes the output of the first layer and performs convolution 
	with $8\times4$ kernels. Both convolutions are performed to retain 
	the input size with the appropriate zero-padding and all kernels are 
	$5\times5$. Max-pooling is 
	used 
	in both
	layers
	to downsample the data, providing feature maps of  $8\times8$ for the 
	first 
	layer and $4\times4$ for the second layer. The 
	activation function is the Rectified
	Linear Unit (ReLu). 
	}}
	\label{fig:neural_network}
\end{figure}

To choose the architecture of the CNN, we used the same training set as in 
\cite{barasuol15iros}. We started from a standard 
architecture (similar to LeNet \cite{lecun1998gradient}) and decreased 
the size of the CNN as much as possible, without compromising the
validation accuracy (which was nearly 100\%).
After increasing the size of the training set, 99\% of the predictions of the 
CNN were feasible footholds (see Equation \eqref{eq:mapping}), from 
which 76\% of the time the prediction was optimal. These 
prediction results are summarized in Section 
\ref{section:cnn_results}. We 
came to
the conclusion that the architecture depicted in Fig. \ref{fig:neural_network}
is suitable for our application. Highly accurate predictions are not needed as
long as the selected foothold is safe according to the heuristics (see 
Table \ref{table:neural_network}).

A comparison between the CNN and the heuristic algorithm in terms of
performance and computational time is also detailed in Section
\ref{section:cnn_results}.
\vspace{-0.3cm}
\subsection{Adjustment of the Foot Swing Trajectory}
After the CNN has been trained, it can infer a foothold adaptation from a
previously unseen heightmap sample. The difference between the optimal
foothold and the nominal one is sent to the trajectory generator module
as a relative displacement to adapt the original foot swing trajectory.

To avoid aggressive control actions, the foothold corrections are collected
from the lift-off to the trajectory apex only. Then, the controller
tracks the latest adjusted trajectory available before the apex.
\section{Results} \label{section:results}
In this section we present the results regarding the computational performance
and the locomotion robustness we achieved with the proposed strategy, both 
in simulation and experiments. 
\vspace{-0.3cm}
\subsection{CNN Prediction Results} \label{section:cnn_results}
We compared the time required to compute a foothold adaptation from the 
same heightmap for both the heuristic algorithm and the CNN. As a metric for
comparison, we use the number of clock tick counts between the beginning
and the end of each computation divide by the computer clock frequency.
To achieve real-time safe performances, these times should be low on average
and display little variance.

To speed up the computation of the heuristics, the algorithm incrementally
expands the search radius from the center of the heightmap, i.e., from the
nominal foothold, and stops once a feasible foothold is found. In the case of
the computation of the full-blown heuristic algorithm, the computational times
range from \SI{0.1}{\milli\second} to \SI{20}{\milli\second}. The prediction
from the CNN takes from \SI{0.072}{\milli \second} to \SI{0.1}{\milli \second}.
The CNN-based model is therefore 15 to 200 times faster than the heuristic
algorithm. Indeed, the computational time increase in a nonlinear fashion 
along with the complexity of the heightmap,
while the neural network presents a computational time less sensitive
to the input. Furthermore, the duration of the control loop 
of our system is \SI{4}{\milli\second}, 
making the computation of the CNN at least 40 times faster than the task rate, 
allowing to run it in real-time.

In Table \ref{table:neural_network},  we summarize the results of the CNN
performance when predicting the optimal footholds. From a total number of 35688
examples (for both front and hind legs) we used half dataset as training set and
half as testing set.
It is
worth noting that the percentage of accurate predictions is not notably high
(about 76\% for both legs). Nevertheless, by looking closely to the false
positives, approximately 99\% of them were sub-optimal decisions according to
the distance criterion, yet deemed as safe with respect to the other criteria 
(see Section
\ref{section:foothold_adaptation}). This means that, if the foothold is not 
optimal, the chosen foothold is still safe.

Regarding the quality of predicted footholds with respect 
to our previous work \cite{barasuol15iros}, we initially compared the results 
of both learning algorithms (CNN vs logistic regression) applied to the same 
training set used to train the logistic regression classifier, consisting of 
3300 examples with 9 possible 
footholds. The output layer of the CNN was initially set to be $9\times 1$, 
matching the number of possible footholds. We improved the prediction accuracy 
to nearly a 100\% using 
the CNN-based classifier, compared to a 90\% using the logistic regression. 
This result and the automation of the training process 
drove us to increase the number of outputs 
(from 9 to 226). We compared the 
CNN-based classifier with a logistic regression using 226 possible outputs 
and a much larger number of examples. The CNN classifier proved to have 
better prediction accuracy (76\% vs 68\%) and yielded a much lower number 
of parameters (8238 vs 51076).
\bgroup
\def\arraystretch{1.25}
\begin{table}[]
	\centering
	\caption{Results of prediction coming from the neural network on the
	test set.}
	\label{table:neural_network}
	\begin{tabular}{|l|l|l|l|}
		\hline
		Leg   & \begin{tabular}[c]{@{}l@{}}Perfect \\ 
		match\end{tabular}         & \begin{tabular}[c]{@{}l@{}}Unsafe \\ 
		footholds\end{tabular} & \begin{tabular}[c]{@{}l@{}}Safe \\ 
		footholds\end{tabular}     \\ \hline
		Front & \begin{tabular}[c]{@{}l@{}}13718/17844 \\ 
		(76.88\%)\end{tabular} & \begin{tabular}[c]{@{}l@{}}47/17844\\  (0.26 
		\%)\end{tabular}   & \begin{tabular}[c]{@{}l@{}}17797/17844\\  (99.74 
		\%)\end{tabular} \\ \hline
		Hind  & \begin{tabular}[c]{@{}l@{}}13700/17844\\ 
		(76.78\%)\end{tabular}  & \begin{tabular}[c]{@{}l@{}}21/17844\\  (0.12 
		\%)\end{tabular}   & \begin{tabular}[c]{@{}l@{}}17823/17844\\  (99.88 
		\%)\end{tabular} \\ \hline
	\end{tabular}
\end{table}
\egroup
\vspace{-0.3cm}
\subsection{Simulation and Experimental Results} 
\label{section:sim_and_exp_results}
To assess improvements in terms of locomotion robustness, we created 
challenging scenarios composed of a series of gaps. These multi-gap terrain
templates are built up from short beams (\SI{15}{\centi\meter} height and
\SI{20}{\centi\meter} width), equally spaced by \SI{10}{\centi\meter}, and
pallets (\SI{15}{\centi\meter} height). 
This scenario is used for both simulation and experimental tests: a 
nine-gap 
template
for the simulated tests (see Fig. \ref{fig:trot_over_gaps}) and a four-gap
template for the experimental ones (see Fig. \ref{fig:hyq_overview}).

In both simulation and experimental tests the locomotion robustness is evaluated
while the robot is performing a trotting gait over the beams at different velocities (\SI{0.3}{\meter\per\second} and \SI{0.5}{\meter\per\second}) and under external
disturbances. The locomotion robustness is evaluated by observing the tracking of
the robot desired velocity and trunk height. To evaluate the performance
repeatability, we considered the data of 5 trials for each desired velocity.
The trotting gait is performed with step frequency of 1.4 Hz, duty factor of 0.65
and a default step height of \SI{12}{\centi \meter}.

We will end the section with complementary experimental results that show the
implementation of our strategy to provide foothold adaptation on a static crawl
algorithm \cite{focchi2016}.
\subsubsection{Simulation Results} \label{section:sim_results}
\begin{figure*}[h!]
	\centering
	\includegraphics[width=0.99\linewidth]{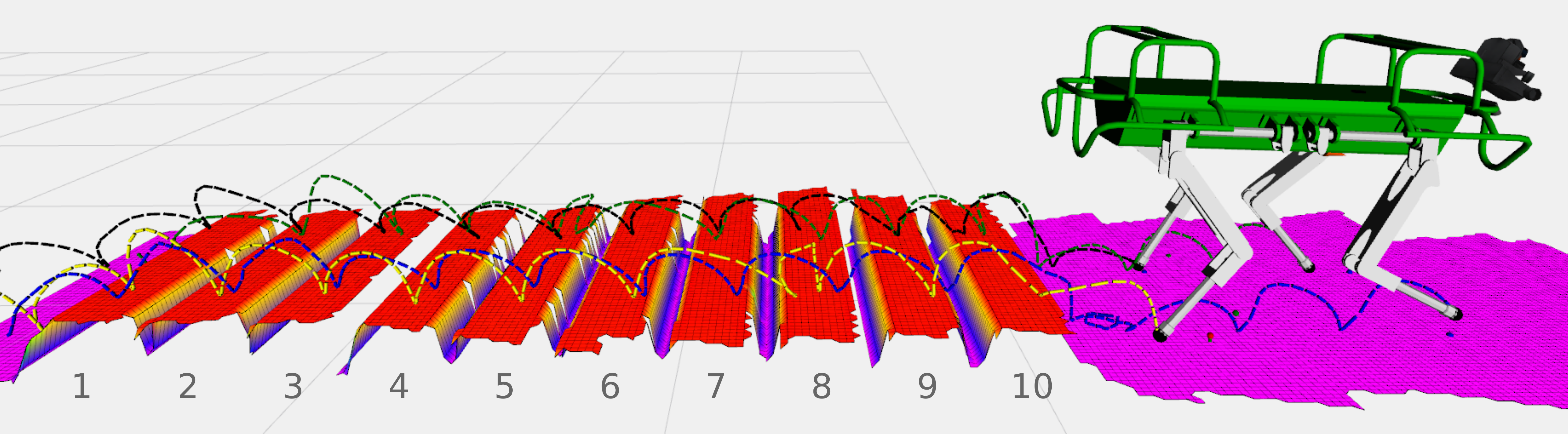}
	\caption{\small{Simulation of gap crossing scenario. 
	Dashed lines correspond to feet trajectories corrected 
	by the
	vision-based adaptation, during \SI{0.5}{\meter\per\second} trotting 
	gait. The 
	feet
	trajectories are identified as: left-front (green), right-front (blue), 
	left-hind
	(black) and right-hind (yellow).}}
	\label{fig:trot_over_gaps}
\end{figure*}
Fig. \ref{fig:trot_over_gaps} depicts the details of the simulation scenario
showing the elevation map computed by the perception system
and the resulting feet trajectories during a multi-gap crossing. Through 
the
footprints (dashed lines) it is possible to see the effect of the foothold
adaptation on the original feet trajectories.

\begin{figure}
	\vspace{-0.7cm}
	\centering
	\includegraphics[width=\linewidth]{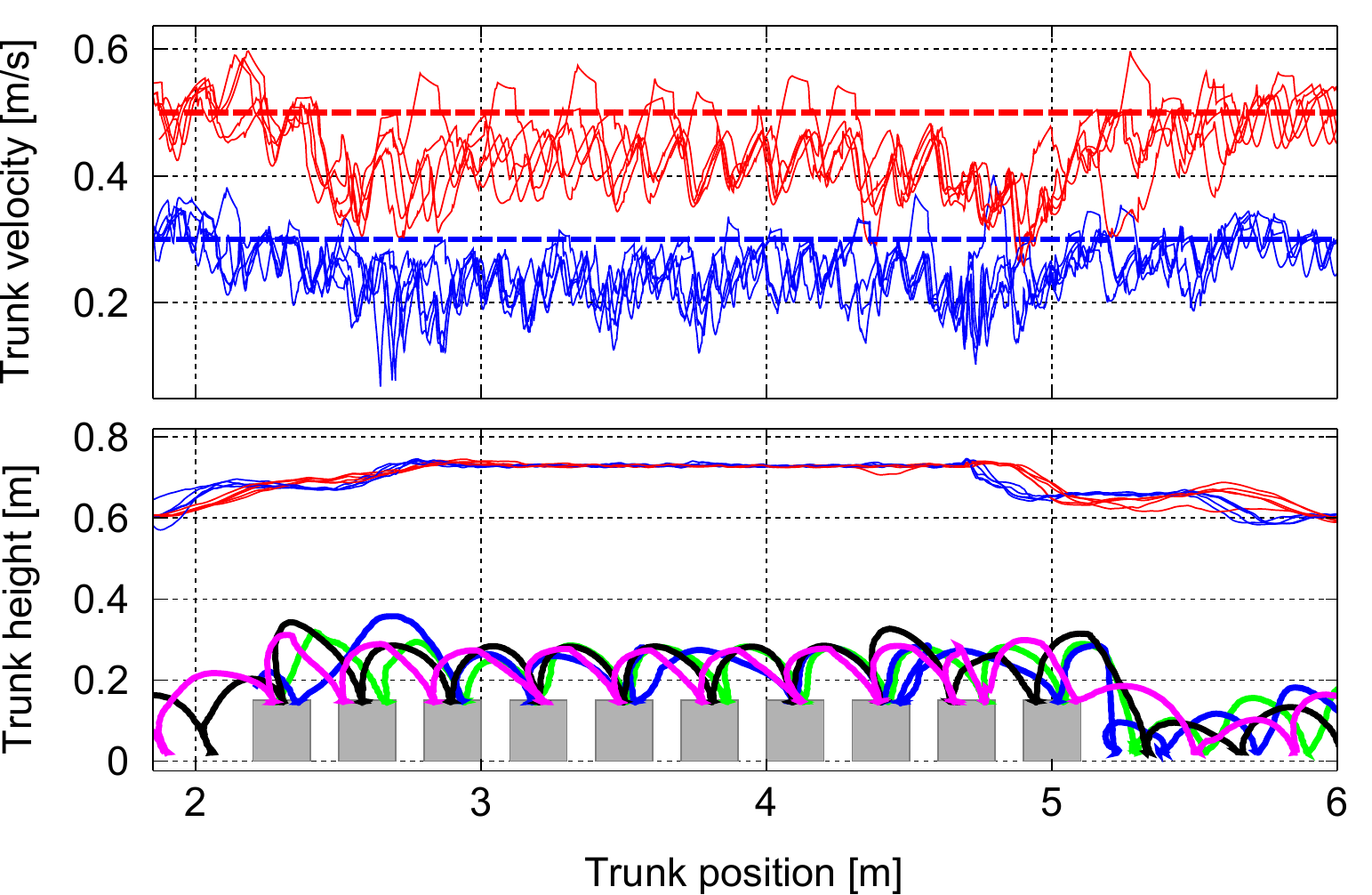}
	\caption{\small{Trotting over multi-gap scenario at different velocites. The
		top figure shows five simulation trials for \SI{0.3}{\meter\per\second} 
		(blue) and
		\SI{0.5}{\meter\per\second} (red) trot. Dashed lines 
		represent desired crossing velocity. The bottom figure shows the trunk
		height while crossing, the position of the beams and one example
		of the resulting feet trajectories for one of the trials 
		performed at
		\SI{0.5}{\meter\per\second}.}} 					 	
	\label{fig:trot_over_gaps_vel}
\end{figure}

\begin{figure}
	\centering
	\includegraphics[width=\linewidth]{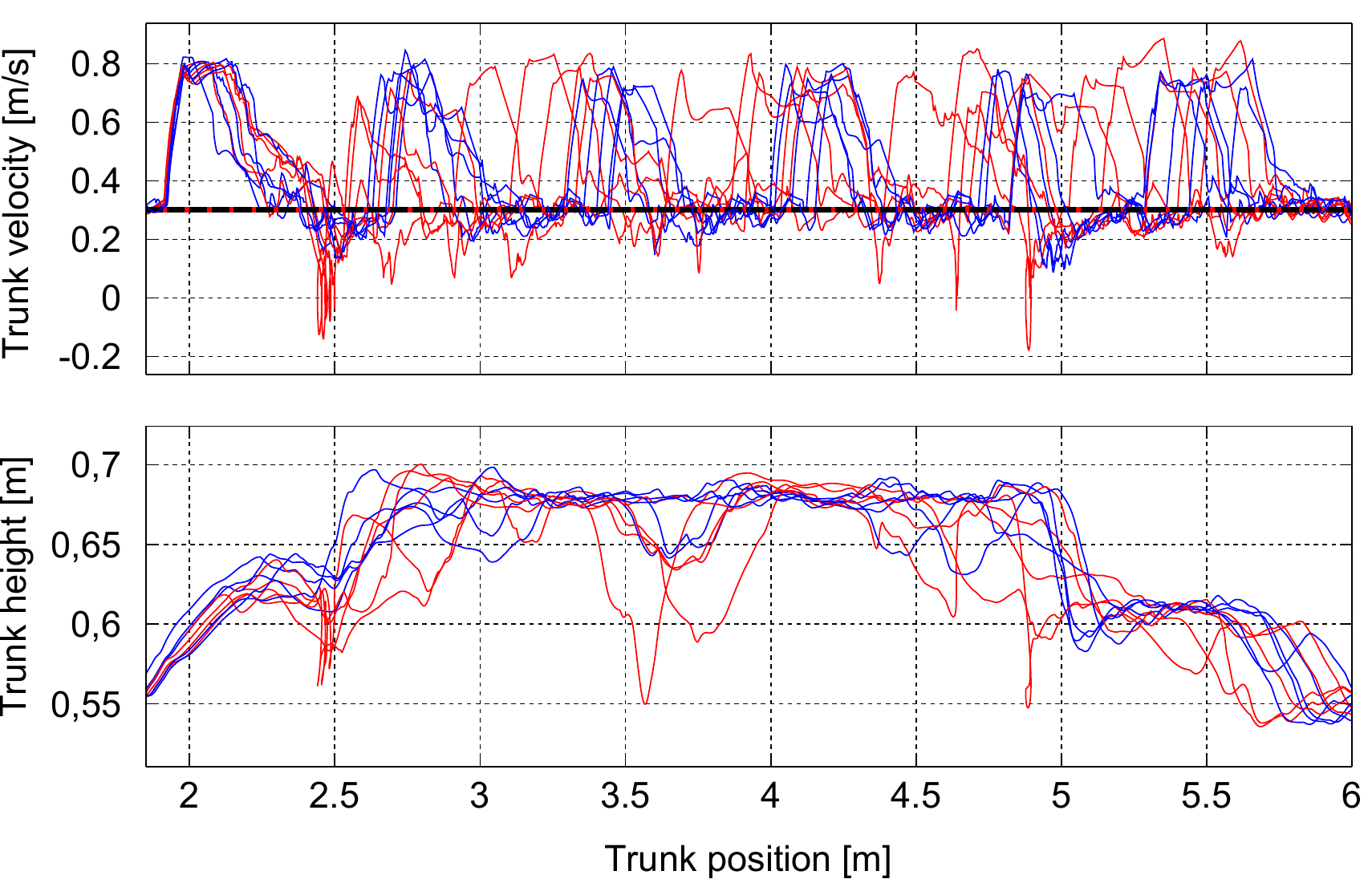}
	\caption{\small{Variation of velocity with 
	\SI{0.1}{\second} force disturbance 
	of \SI{500}{\N} every \SI{2}{\second} and the trunk height while 
	crossing multiple gaps for several simulation
	trials. The blue lines represent the trials performed with the 
	correction set to be executed continuously during swing.
	The red lines show the trials where the correction is 
	computed only once at the lift-off for each leg.}}
	\label{fig:trot_over_gaps_push}
\end{figure}

Using the beam numbers shown at the bottom of Fig. \ref{fig:trot_over_gaps},
several examples of corrections that avoided stepping inside the gap can be 
easily
identified: left-front foot double stepping on beam 5 and stepping over beam 6
(green line); right-hind foot stepping over beam 7 and double stepping on beam 8
(yellow line). The right-front foot steps over beam 10 (blue line) to avoid
placing the foot too close to the beam edge. The results of the 5 trials of the
multi-gap crossing at different velocities are shown in Fig.
\ref{fig:trot_over_gaps_vel}.
As a last simulation example, we test the capabilities of our strategy to respond
against external perturbations and show the benefits of having a continuous
adaptation. During the same gap-crossing task shown before, with a \SI{0.3}{\meter\per\second} trotting, we apply a series of
perturbations of \SI{500}{\newton} every \SI{2}{\second} with a duration of
\SI{0.1}{\second} each. The perturbations are applied on the base longitudinal
direction disturbing the forward motion. Two cases are studied in this setup: the
first corresponds to the case when the adaptation is only computed at the
lift-off, and the trajectory is not corrected during the swing phase (red lines in
Fig. \ref{fig:trot_over_gaps_push}); in the second case, the foothold adaptation
is continuously computed and can be modified along the swing phase (blue lines
in Fig. \ref{fig:trot_over_gaps_push}). It can be seen that in some cases, when
the adaptation is only computed at the lift-off, the velocity of the trunk
decreases considerably and the trunk height is less stable. Such tracking errors,
caused by undesired impacts with the beams, happen due to the lack of foothold 
adaptation during the swing phase.
\subsubsection{Experimental results}\label{section:exp_results}
\begin{figure} [h!]
	\vspace{-0.2cm}
	\centering
	\includegraphics[width=\linewidth]{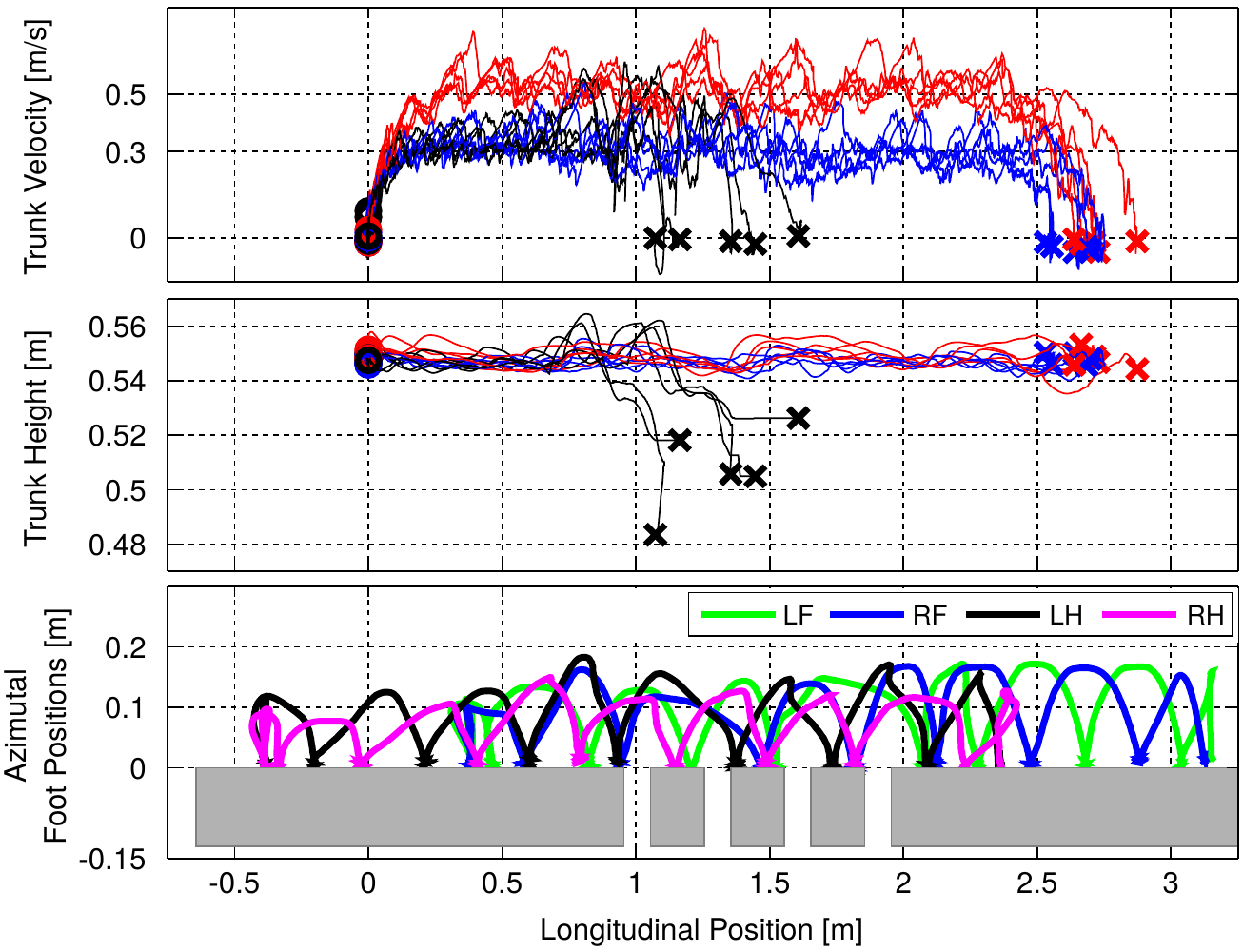}
	\caption{\small{Experimental trotting over gaps, for different desired trunk
		forward velocities, with and without visual feedback. The two top 
		plots correspond to trials at \SI{0.3}{\meter \per \second} 
		(blue), \SI{0.5}{\meter \per \second} (red) with visual feedback. Black 
		lines show trials at \SI{0.3}{\meter
				\per \second} without visual feedback. 
		Five trials are shown for each of the
		cases. Circle and cross markers indicate the beginning and the end, respectively,
		of each corresponding trajectory. The bottom plot shows the position of the
		beams and the resulting feet trajectories for one of the trials 
		performed at
		\SI{0.5}{\meter\per\second}.}}
	\label{fig:trot_gap_exp}
\end{figure}

we prepared the setup depicted in Fig.
\ref{fig:hyq_overview} for the gap-crossing 
experiments. To show how challenging 
this task
can be for a blind robot, we also present the results of 5 
trials
using the RCF without the foothold adaptation. The results of these
trials are shown in Fig. \ref{fig:trot_gap_exp}.

\begin{figure} [h]
	\centering
	\includegraphics[width=\linewidth]{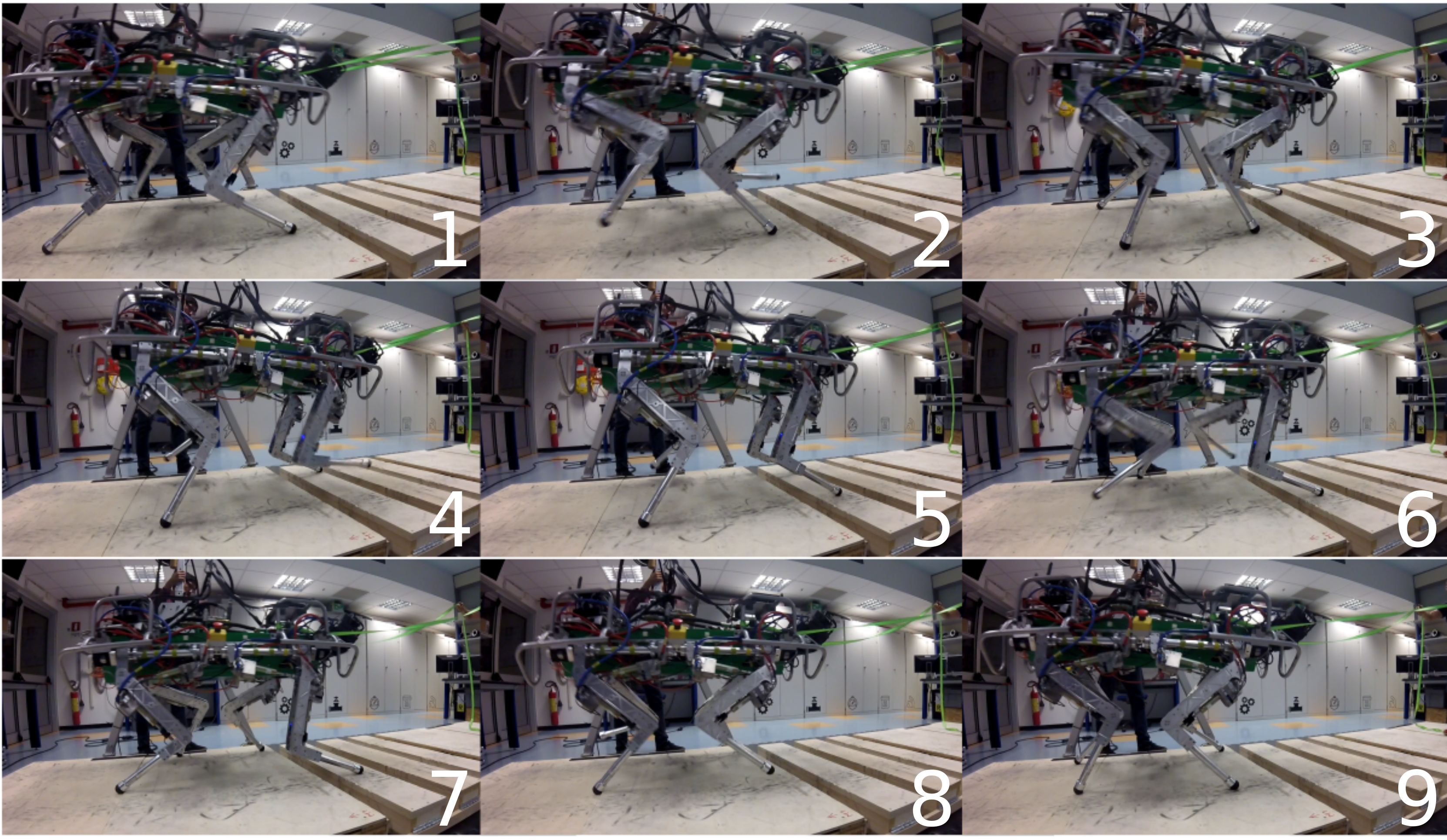}
	\caption{\small{Video screenshots of the disturbance test showing the robot
		reactions while being pulled towards the gaps.}}
	\label{fig:dist_over_gaps}
\end{figure}

\begin{figure}
	\centering
	\includegraphics[width=\linewidth]{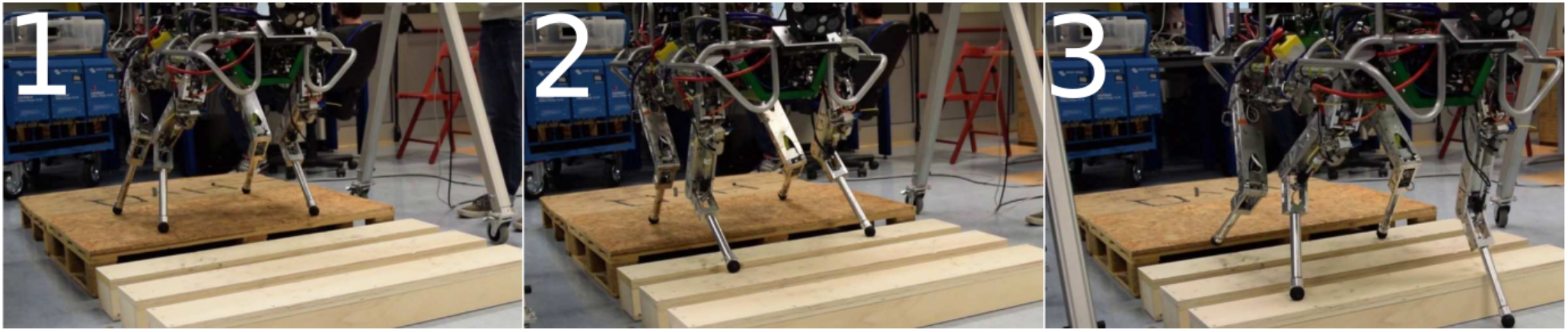}
	\caption{\small{Video screenshots of a crawl experiment showing the 
	robot crossing a series of gaps. 
}}
		\label{fig:experiment_picture}
		\vspace{-0.15cm}
\end{figure}

As it can be seen, the robot was not able to complete the task without the
visual-based adaptation. 
With visual foothold adaptation the goal was achieved with similar 
performances 
between the \SI{0.3}{\meter\per\second} and \SI{0.5}{\meter\per\second} trials. 
The
resulting feet trajectories of one of the trials at 
\SI{0.5}{\meter\per\second} can
also be seen at the bottom part of Fig. \ref{fig:trot_gap_exp}. 
The robustness of the robot against external disturbances was also experimentally
tested on top of the multi-gap terrain template. For such test, we placed
the robot on top of one of the pallets, displayed on the bottom-left in Fig.
\ref{fig:dist_over_gaps}, and commanded the robot to keep its position on it while
trotting.
Then, we disturbed the robot
by pulling it and forcing it to go repeatedly over the gaps. Fig. 
\ref{fig:dist_over_gaps}
shows video screenshots of disturbance experiments where the robot is subject to
strong pulling forces (estimated to be around \SI{500}{\newton}). It can also 
be observed
that the robot is able to keep its balance and to come back to its commanded position
without stepping inside the first gap.
It is worth highlighting the robust autonomy of the closed-loop system, since 
the
robot is only commanded to keep its global position and no base trajectory is
pre-designed while it is disturbed.
As complementary result, we show the generality of the proposed strategy by
implementing it into a blind static crawl algorithm \cite{focchi2016}.
In the case of the crawl, we created a gap-crossing scenario where the robot has
to step on a series of wooden beams to then take a step down and reach the
ground. For this task, the robot is commanded to go forward with a velocity of
\SI{0.1}{\meter\per\second}. We compare the results between the haptic
blind crawling strategy and enhanced with the CNN-based foothold 
adaptation.
In the case of a static crawl the nominal footholds are already provided in
the world frame and do not need to be predicted. Therefore, we compute the
correction only at lift-off and execute the trajectory without changing it
during the swing phase. It can be seen in the series of snapshots in Fig.
\ref{fig:experiment_picture}, and in the attached video, that the robot 
succeeded in
crossing the 
scenario when the CNN-based foothold adaptation is 
implemented.
The robot was not able to cross the gaps without foothold adaptation.
\section{Conclusions} \label{section:conclusions}
We have presented a novel strategy for continuous foothold
adaptation based on a Convolutional Neural Network. We evaluated the
performance of the approach by performing dynamic trotting and static
crawling gaits on a challenging surface, 
difficult to traverse
if only proprioceptive sensors and haptics are used. The various simulated
and experimental 
trials, at different forward velocities and
under external disturbances, demonstrated the robustness of the strategy
and its 
repeatability. Moreover, we showed that the proposed
strategy is more robust with respect to the ones that only adapt the
nominal foothold at the leg lift-off.

The CNN resulted to be up to 200 times faster than computing the full-blown
heuristics to find a safe foothold, showing that the strategy has
potential to deal with more complex heuristics and still satisfy
the real-time constraints.
Due to 
the low computational load of the method (40 times faster than the task rate), 
our future work will concentrate on
learning more complex heuristics that evaluate footholds in a two-step
horizon, dynamic criteria for better robot balancing, posture adjustment
and gait parameter modulation. Moreover,  
we will customize even further the CNN architecture to better reflect the 
computations performed by the heuristics. 
\section*{Acknowledgements}	
We would like to thank the following members of the DLS lab for the help 
provided while 
writing this paper: Geoff 
Fink, Andreea 
Radulescu, Shamel 
Fahmi, Lidia Furno, Andrzej Reinke, Evelyn D'Elia, Fabrizio Romanelli, Gennaro Raiola, Jonathan Brooks and 
Marco Ronchi.


\bibliographystyle{IEEEtran}
\bibliography{references}
\end{document}